\documentclass[sigconf,manuscript,nonacm]{acmart}

\makeatletter
\def\@ACM@checkaffil{
    \if@ACM@instpresent\else
    \ClassWarningNoLine{\@classname}{No institution present for an affiliation}%
    \fi
    \if@ACM@citypresent\else
    \ClassWarningNoLine{\@classname}{No city present for an affiliation}%
    \fi
    \if@ACM@countrypresent\else
        \ClassWarningNoLine{\@classname}{No country present for an affiliation}%
    \fi
}
\makeatother

\AtBeginDocument{%
  }

\begin{document}

\title[Fairness of Deep Ensembles: On the interplay between per-group task difficulty and under-representation]{Fairness of Deep Ensembles: \\On the interplay between per-group task difficulty and under-representation}

\author{Estanislao Claucich}
\email{eclaucich@sinc.unl.edu.ar}
\affiliation{%
  \institution{sinc(i), CONICET \& Universidad Nacional del Litoral}
  \city{Santa Fe}
  \state{Santa Fe}
  \country{Argentina}
}

\author{Sara Hooker}
\affiliation{%
  \institution{Cohere For AI}
  }
\email{sarahooker@cohere.com}

\author{Diego H. Milone}
\email{dmilone@sinc.unl.edu.ar}
\affiliation{%
  \institution{sinc(i), CONICET \& Universidad Nacional del Litoral}
  \city{Santa Fe}
  \state{Santa Fe}
  \country{Argentina}
}

\author{Enzo Ferrante}
\authornote{Equal contributtion.}
\email{eferrante@sinc.unl.edu.ar}
\affiliation{%
  \institution{Institute of Computer Sciences, CONICET \& Universidad de Buenos Aires}
  \city{Buenos Aires}
  \state{Buenos Aires}
  \country{Argentina}
}

\author{Rodrigo Echeveste}
\authornotemark[1]
\email{recheveste@sinc.unl.edu.ar}
\affiliation{%
  \institution{sinc(i), CONICET \& Universidad Nacional del Litoral}
  \city{Santa Fe}
  \state{Santa Fe}
  \country{Argentina}
}


\begin{abstract}
Ensembling is commonly regarded as an effective way to improve the general performance of models in machine learning, while also increasing the robustness of predictions. When it comes to algorithmic fairness, heterogeneous ensembles, composed of multiple model types, have been employed to mitigate biases in terms of demographic attributes such as sex, age or ethnicity. Moreover, recent work has shown how in multi-class problems even simple homogeneous ensembles may favor performance of the worst-performing target classes. While homogeneous ensembles are simpler to implement in practice, it is not yet clear whether their benefits translate to groups defined not in terms of their target class, but in terms of demographic or protected attributes, hence improving fairness. In this work we show how this simple and straightforward method is indeed able to mitigate disparities, particularly benefiting under-performing subgroups. Interestingly, this can be achieved without sacrificing overall performance, which is a common trade-off observed in bias mitigation strategies. Moreover,  we analyzed the interplay between two factors which may result in biases: sub-group under-representation and the inherent difficulty of the task for each group. These results revealed that, contrary to popular assumptions, having balanced datasets may be suboptimal if the task difficulty varies between subgroups. Indeed, we found that a perfectly balanced dataset may hurt both the overall performance and the gap between groups. This highlights the importance of considering the interaction between multiple forces at play in fairness.
\end{abstract}

\maketitle

\section{Introduction}
Model ensembling \cite{Breiman2001,Dietterich2000} is a widespread technique to boost the performance of machine learning (ML) models in general, and deep neural networks (DNNs) in particular. When it comes to fairness, ensembling techniques pooling multiple model types (heterogeneous ensembles) have been proposed in order to mitigate biases \cite{kenfack2021, chen2017, gohar2023}. These methods require however training different architectures, complicating their applicability and the interpretability of the ensemble model. Moreover, even simple homogeneous ensembles sharing the same architecture and hyper-parameters have recently shown to reduce performance disparities in terms of the target class in multi-class problems \cite{ko2023fair}. However, the work by Ko et al. \cite{ko2023fair} limits its scope to understanding the impact of ensembles on a target class (class label) in a supervised problem setting. Often, issues with fairness are concerned with performance on a protected attribute which defines sub-populations of the distribution being modeled \cite{caton2024fairness}. Here we explore whether this result would also apply to subgroups defined not in terms of their target class, but in terms of protected attributes. That is, we considered the effectiveness of ensembles as a bias mitigation strategy when performing sub-population analysis.

Under-representation of protected sub-groups is a known cause for biases in ML models, whereby model performance tends to be lower for minorities or minoritized groups \cite{Larrazabal2020,buolamwini2018gender,yang2020towards,hooker2020characterisingbiascompressedmodels}. A common and effective bias mitigation strategy is then to rebalance the data in terms of protected attributes \cite{ricci2022addressing}. However, in most cases, the gap is reduced by gaining performance for one group at the expense of losing performance in another one (known as the leveling-down effect) \cite{zietlow2022}. This trade-off creates an ethical dilemma and, in critical domains such as healthcare, mitigating biases in such a way may go against basic principles of bioethics such as that of non-maleficence \cite{beauchamp2003methods}. Hence, it is an increasingly urgent direction of research to find solutions which are able to reduce this gap without negatively impacting the performance of any subgroup. 

Moreover, balance in terms of protected attributes may interact with other factors, such as a varying degree of task difficulty between sub-groups \cite{ferrante2024open,Agarwal_2022_CVPR,siddiqui2023metadata}. Indeed, in many scenarios (e.g. medical imaging) differences related to sex, gender, skin-tones or ethnicity, among others, may lead to varying levels of task difficulty for different demographic sub-groups.
In practice, however, the intrinsic difficulty of the task for different subgroups is not trivial to determine and. As noted in \cite{weng2023}, it can be difficult to explain where the performance gaps originate for certain groups or datasets. In this work we took the performance of models trained on a balanced dataset as a proxy for relative difficulty between sub-groups. This allowed us to explore the question of how ensembles may leverage different and sometimes conflicting sources of bias. 

In all, our work showcases the potential of homogeneous ensembles to mitigate performance gaps both due to under-representation and uneven task difficulty. Critically, we show how this may be achieved without a loss in overall performance. Across the different experiments, the relative improvements obtained with homogeneous ensembles ranged from 1.3\% to 4\%, reducing the gap between the groups to be insignificant. We believe this work opens up a valuable avenue of research for the generalization to other domains and tasks, with potentially ample implications for the fairness community.

\section{Related work}
Although deep ensembles have been proposed as a means to improve model performance \cite{lakshminarayanan2017simple}, most existing research focuses on the impact of ensemble diversity on model accuracy and robustness \cite{lee2015m,pividori2016,nam2021diversity,larrazabal2021orthogonal}, while the implications in terms of fairness of such models remain largely unexplored. The recent work by Ko et al. \cite{ko2023fair} is among the first to discuss these implications for deep ensembles. Their study postulates that fairness naturally emerges from deep ensembling, as the performance of the disadvantaged group improves disproportionately with an increasing number of models compared to the advantaged group. However, when performing sub-group analysis to identify advantaged and disadvantaged groups, the authors focus on target classes rather than following the standard practice in the fairness community of analyzing group fairness by protected attributes—i.e. \cite{Seyyed-Kalantari2021,seyyedkalantari2020chexclusionfairnessgapsdeep}, characteristics of the population such as gender or age, which are not the target labels for classification.

The closest work to ours is that of Schweighofer et al. \cite{schweighofer2024disparate}, which, similarly to our study, adheres to the standard group fairness framework and evaluates fairness metrics by considering sub-populations characterized by protected attributes. Their findings indicate that deep ensembles unevenly benefit different protected groups (referred to as the \textit{disparate benefit effect}). However, their focus is on how this disparate benefit relates to the predictive diversity of the ensemble, that is, on how varied the predictive distributions of different models of the ensemble are. They argue that it is because of the difference in predictive distributions between members of the ensemble, that the group can improve on individual models. Consistent with other research demonstrating that predictive diversity can serve as a proxy for anticipating biases in deep neural networks \cite{mansillademographically,hooker2020characterisingbiascompressedmodels}, they show that tasks exhibiting this disparate benefit effect display significant differences in average predictive diversity between groups, while tasks without this effect exhibit only minimal differences.
In this work, we take a different approach by investigating a factor overlooked in previous studies: the impact of task difficulty on this disparate effect and how it relates to the over- or under-representation of disadvantaged populations in bias mitigation efforts.

\section{Evaluating fairness of ensembles under multiple sources of bias}

To assess the potential of homogeneous ensembles to mitigate biases and study the interaction between the under-representation of a certain group and its corresponding task difficulty, we considered synthetic and real scenarios.

\subsection{Synthetic scenarios}
We begun by building two synthetic scenarios that allowed us to control every aspect of the experiment. Both of them consist of a simple binary classification task using fully connected (FC) layers. In each case, the samples are generated from Gaussian distributions representing two different demographic subgroups and two target classes. In order to build an intuition of how this translates to a typical fairness scenario in the healthcare domain, we define these subgroups as male (M) and female (F), and the target class as healthy (0) or diseased (1). We acknowledge these groupings are overly simplistic, but serve as a toy example to first explore the phenomena in a fully controlled way.

The distributions were constructed with the same standard deviation ($\sigma=0.2$), and their centers were symmetrically positioned with respect to the origin, to discard any asymmetries originating from weight regularization terms in the cost function. Datasets are sampled from these distributions and the degree of balance or under-representation can be controlled by the number of samples taken from each. We took different ablations of the representation for each subgroup raging from 0\%-100\% to 100\%-0\% for male-female, with a step of 10\%. In all cases, datasets are balanced in terms of target class.

The two synthetic scenarios differ only in how we manipulated task difficulty for the groups. Namely, when a model trained with a perfectly balanced dataset performs equally well for every group, we consider the task difficulty to be the same for each of those groups. On the other hand, if performance is lower for a sub-group in a balanced setting, we take the task difficulty to be higher for that group. We follow two different strategies to increase task difficulty:

\begin{enumerate}
    \item \textit{Target class label noise}: Increased difficulty for the female sub-group is induced by having a varying percentage of randomly selected samples with their target class labels flipped (see Figure \ref{fig_impact_ensembles}a). We tested scenarios with a label noise ranging from 0\% to 50\% with a step of 5\%.
    
    \item \textit{Rotating decision boundary}: Increased difficulty arises from the miss-alignment of optimal decision boundaries for both groups. The optimal decision boundary for the female subgroup was rotated with respect to the horizontal line (see Appendix A for an example). The use of an L1 regularization when training the models generates an increased difficulty for the rotated subgroup, breaking the symmetry of the problem. We rotated the boundary from 0 to 45 degrees, with a step of 5.
\end{enumerate}

Without loss of generality, we vary the difficulty of the task for the female group while the male subgroup has a constant difficulty.

\subsection{Real scenarios}

While we adopted a basic FC architecture for the synthetic case, we considered more complex architectures for the real datasets. In this case, the models consisted of ResNet-50 networks \citep{he2015deepresiduallearningimage} pre-trained on Image-Net, and fine-tuned for the CelebA \cite{celebA} and CheXpert \cite{chexpert} datasets. In each case the male and female subgroups were constructed in terms of the gender attribute in CelebA and sex attribute in CheXpert, defined for each sample. 

\textit{CelebA} consists of images of celebrities with multiple attributes. In this case we tackled a binary hair-color prediction task with labels blond (B) and not-blond (NB) as defined in prior works \citep{sagawa2020investigationoverparameterizationexacerbatesspurious,hooker2020characterisingbiascompressedmodels}. In CelebA, to avoid known spurious correlations, we balanced the dataset not only for male and female subgroups but also for the blond attribute. This means that the four possible groups (M-B, M-NB, F-B, F-NB) have the same number of samples ($N=4,350$) in our pre-processed dataset. 

\textit{CheXpert} consists of frontal and lateral views of chest radiographs for different patients. In this case, a binary classification for lung opacity was performed. This class was created by fusing the labels edema, consolidation, lung opacity, pneumonia, atelectasis, and lung lesion into a single class, as they manifest similar findings in the image.  As CheXpert has several samples for a single patient, we first reduced the dataset by selecting only one sample per patient. The final dataset was balanced by gender and opacity attributes, once again having the same number of samples for the four possible subgroups. 

With these setups, we once again varied the balance ratio of male and female samples during training. We did not however modify the relative difficulty of the task as we did for the synthetic scenarios, but rather inherited it by the problem itself. In these cases, we define the most difficult group as the one that under-performs in a balanced setting, i.e. for a 50\%-50\% representation of male-female. 

\section{Results}

\subsection{Synthetic Scenarios}

For each ablation of per-group representation and task difficulty in both synthetic scenarios we trained 20 models differing only in their initialization. Moreover, 5 different folds were created for statistical purposes. With these models, we generated ensembles of varying sizes by randomly selecting the corresponding number of models (from 1, i.e. a single model, to an ensemble of 20 models). We did 500 random selections of the models to avoid the results being susceptible to the order in which the models are selected, averaging the results for all selections. Each model of an ensemble is trained independently and with the same training samples. This is important to guarantee that the amount of observed data does not change with the ensemble size. With this homogeneous ensemble setting the only difference between the models is the stochasticity during initialization and training.

\begin{figure}[]
\centering
\includegraphics[width=1\textwidth]{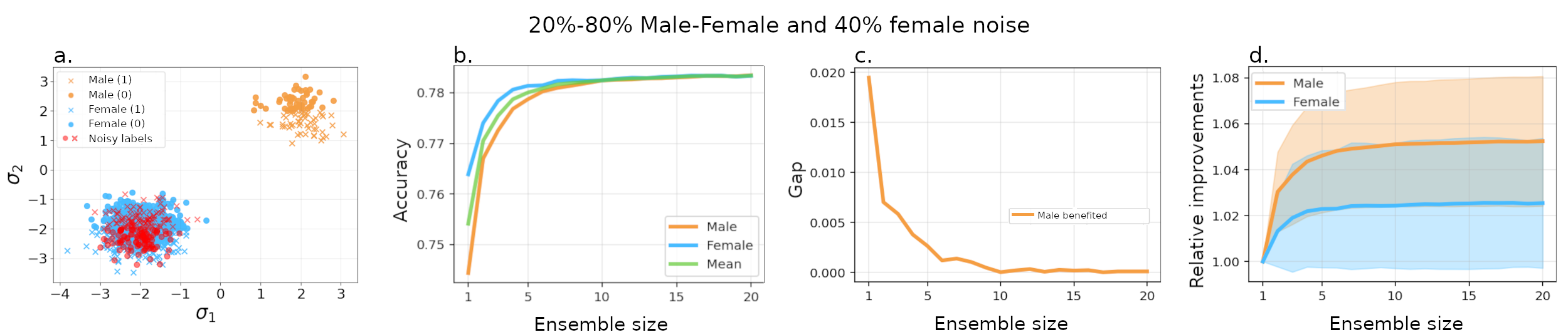}
\caption{Impact of ensembling on an imbalanced scenario (20\% male and 80\% female) and with a label noise applied to 40\% of female samples before training the models. Ensembles help to improve the performance of both subgroups, translating into an improvement of the overall performance. The relative improvements tend to be greater for the underperforming one (male in this scenario), reducing the gap between the groups.}\label{fig_impact_ensembles}
\end{figure}

In Figure \ref{fig_impact_ensembles} we take an illustrative example for the \textit{label noise} scenario, for a 20-80 balance ratio for M-F and 40\% of flipped labels for the F subset (Fig. \ref{fig_impact_ensembles}a). We first observe that the performance of the model in terms of accuracy improves as the number of models in the ensembles grow, both in average (from 75.3\% to 78.3\%) and for each population (Fig. \ref{fig_impact_ensembles}b). For example, we see maximum gains of 5\% for M sub-group. Notably, the gap between the best performing sub-group and the worst is reduced, from 1.9 points of accuracy to marginally zero (Fig. \ref{fig_impact_ensembles}c). These overall gains mean that homogeneous ensembles improve fairness and performance for every group simultaneously, as we further discuss below. We also compute the relative improvement for each group (Fig. \ref{fig_impact_ensembles}d), which shows the same tendency as previously reported in \citet{ko2023fair} for target classes, now for protected attributes.

\begin{figure*}[]
\centering
\includegraphics[width=1\textwidth]{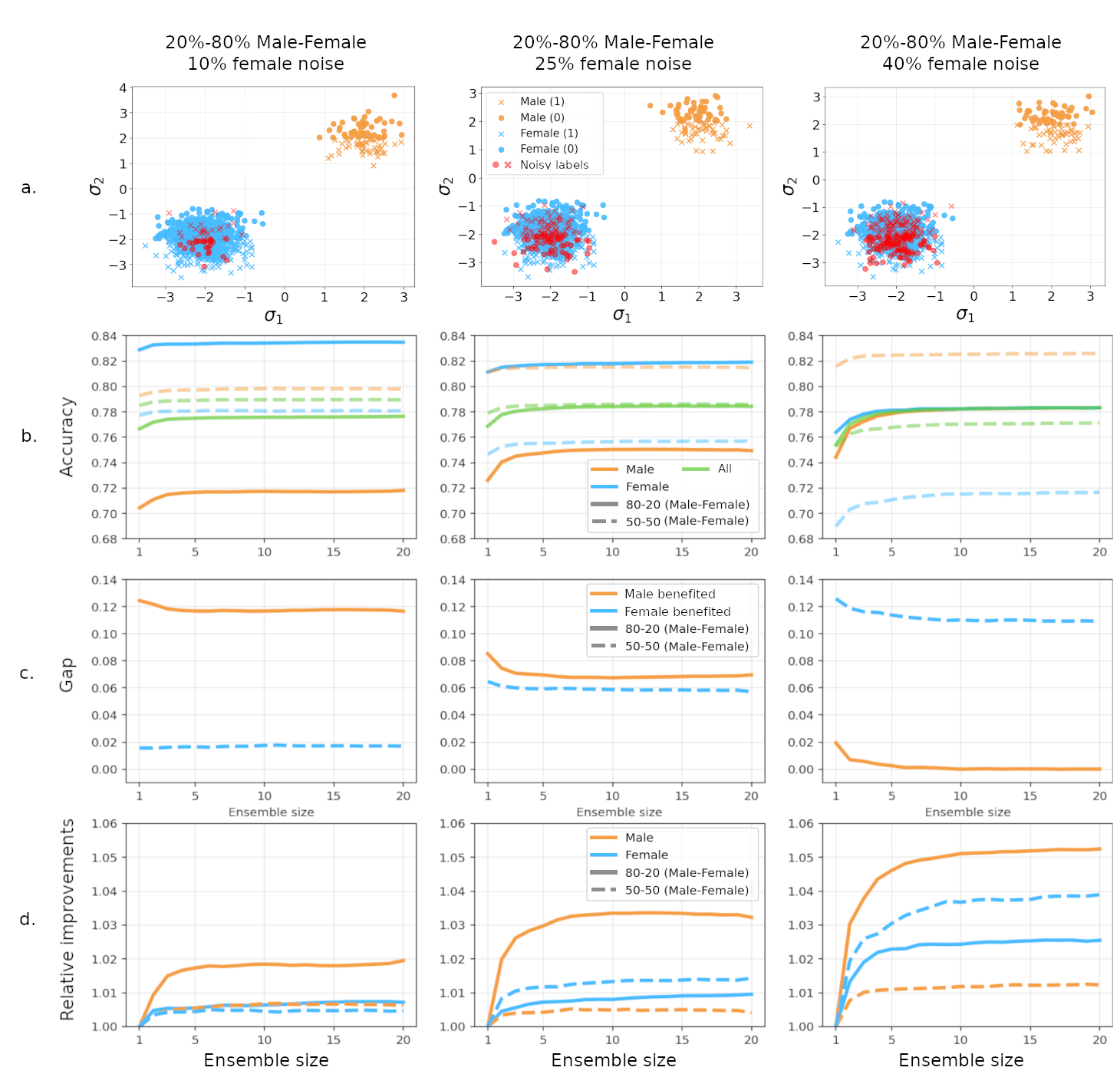}
\caption{Results for the \textit{label noise} synthetic scenario. Each column represents different noise levels in the female subgroup while keeping a constant 20\% under-representation of males  (20-80 M-F). (a) Samples of both distributions under different noise-balance configurations. The following rows correspond to (b) model accuracy for male (orange), female (light blue), and overall (green); (c) absolute performance gap (line color represents the group that is being benefited), and (d) relative improvements by the ensemble size. Dashed lines correspond to the same noise configuration but in a balanced scenario (50-50 M-F).}\label{fig_varying_difficulty}
\end{figure*}

We then studied the tension between per-group difficulty and under-representation. Without loss of generality, we increased the difficulty in the F group, and took an under-representation of the M group. The problem is symmetric and the M and F groups could have been inverted with the exact same results. We begin with a low label-noise level, where the representation dominates, having a higher performance for the F group (Figure \ref{fig_varying_difficulty}, left column), and progressively increasing the difficulty of the F group until these factors compensate and the performance gap considerably narrows down (Figure \ref{fig_varying_difficulty}, see a and b panels, left to right).\\

\begin{figure*}[]
\centering
\includegraphics[width=1\textwidth]{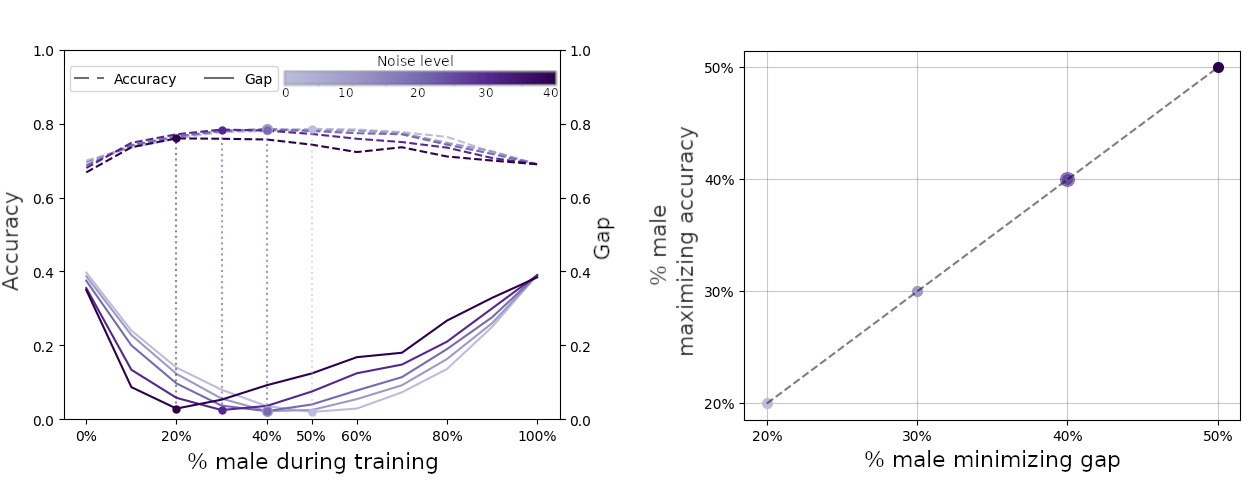}
\caption{Results for the \textit{label noise} synthetic scenario. (Left) Overall accuracy and gap for different noise levels by the balance ratio between male and female subgroups for \textit{label noise} scenario. (Right) Ideal male-female balance ratio for the different noise percentages. As the discrepancy in difficulty increases for both subgroups, more samples are needed for the most difficult one to achieve the best overall performance and minimum gap between the groups.}\label{fig_optimality}
\end{figure*}

\noindent \textbf{Homogeneous ensembles promote positive-sum fairness.} A first important finding of our synthetic analysis is that no matter the noise level or the balance ratio used, ensembles decrease the gap in performance while increasing the individual performance of both groups, constituting a case of \emph{positive-sum} fairness \cite{belhadj2024positive}. That holds for a fixed relative difficulty (which is determined by the task and the nature of the data) and given a fixed imbalance which may come with the dataset. The counterpart of this concept is the \emph{leveling-down} effect, which denotes the tendency for a trade-off between fairness and overall performance, which should be avoided \cite{zietlow2022}. In this sense, our experiments show how homogeneous ensembles induce positive-sum fairness gains, avoiding the leveling-down effect.\\

\noindent \textbf{Subgroups with more difficult tasks benefit more from homogeneous ensembling.} Another interesting behavior was that the magnitude of the relative improvements increased with the difficulty of the task (Figure \ref{fig_varying_difficulty}d). As a reference, the same experiments were conducted with balanced datasets, where only the difficulty varies (see dashed lines in Figure \ref{fig_varying_difficulty}, b to d). We observed that the relative improvement for the under-served population also grew as the relative task difficulty increased in the balanced scenario.\\ 

\noindent \textbf{Bias mitigation requires over-representation under unequal task difficulty.} Interestingly, when looking at the absolute gap we noticed that while the gap was lower for the balanced scenario where both groups had similar difficulty (Fig. \ref{fig_varying_difficulty}c, left), this tendency eventually reverses as one of the tasks becomes more difficult for one of the groups (from left to right). Specifically, for the label flipping variant we observe that at 25\% and 40\% label flipping the gap from the gains from ensembling widen from 1.3 points of accuracy (favoring the balanced scenario) to 10.6 (favoring the imbalanced scenario), middle and right columns at Figure \ref{fig_varying_difficulty}c, respectively. Namely, if a task is harder for one subgroup, achieving a less biased model requires breaking the balance in favor of the harder subset. Similar tendencies in the \textit{rotating decision boundary} scenario can be found in Appendix A. In Figure \ref{fig_optimality} we zoom in into this behavior by plotting the overall accuracy and gap, as a function of M-F imbalance for different noise percentages (see left panel). Starting from a scenario where both groups have the same difficulty level (0\% female noise, less intense line), we found that the ideal balance ratio is 50\%, as expected. And as we increase the noise percentage (i.e. the difficulty) on females, we would need more samples of this subgroup to compensate for this, so the ideal balance ratio will be one with more females than males. Notably, we observe that the M-F ratio that minimizes the performance gap, also maximizes the overall accuracy (Figure \ref{fig_optimality} vertical lines in left panel, and right panel).

\begin{figure}
\centering
\includegraphics[width=1\columnwidth]{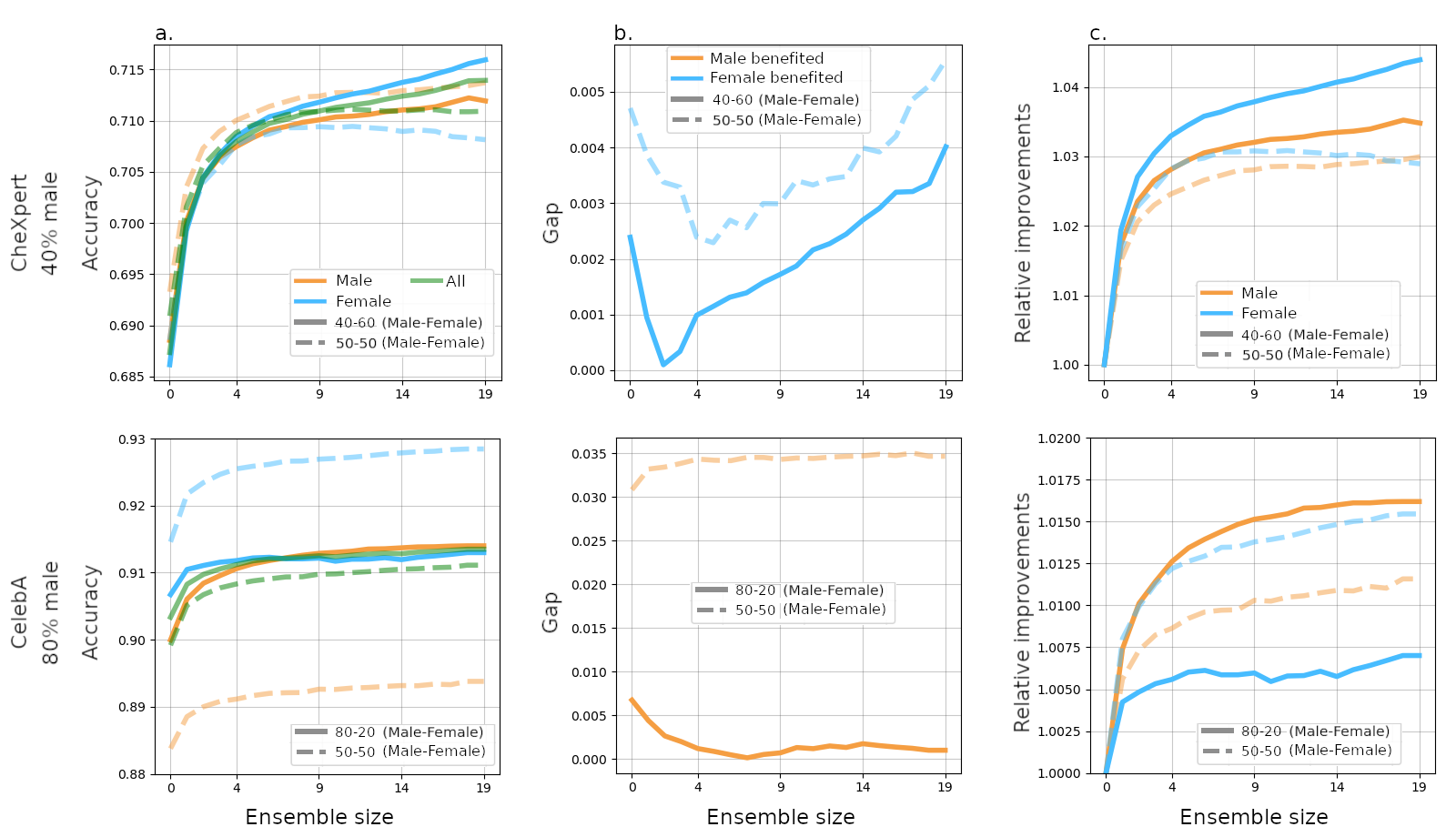}
\caption{(Top) For CheXpert, female was identified as the most difficult group, and we found an ideal balance ratio of 40-60 M-F (more females during training) achieving positive-sum fairness. (Bottom) For CelebA, the male subgroup was the most difficult one, and the ideal balanced ratio was 80-20 M-F (more males during training).}\label{fig_chex_celeb}
\end{figure}

\subsection{Real scenarios: CelebA and CheXpert}

We now see if our findings in a synthetic setting extend to more complex heavily used datasets. For the real world datasets, we followed a similar methodology, where we train 20 models for 5 folds and built ensembles of different sizes. Results for these models follow a similar pattern as observed with the synthetic experiments (Figure \ref{fig_chex_celeb}). Crucially, for both datasets (top and bottom rows) the accuracy systematically increased with the ensemble size for all groups. 

In CheXpert (Figure \ref{fig_chex_celeb}, top row), we saw that by using a balanced dataset the performance difference between males and females was not very significant. Nonetheless, we could also find a non-balanced scenario that achieves positive-sum fairness. In this case, this ratio was close to the balanced scenario (40\%). With this M-F ratio fixed, the under-served population for a single model is F, which is the one with the largest improvement as the ensemble size grows (4.4\% with 20 models in the ensemble). Curiously, the improvement is so big for this group that it ends up having a better performance for larger ensembles. This is evidenced in the v-shaped function for the gap (which is measured as an absolute value). In this case, an ensemble of 3 total models results in the minimal gap, but is important to notice that regardless the ensemble size, the imbalance scenario showcased here achieves a lower gap than the balanced one (see Figure \ref{fig_chex_celeb}b top row).

In CelebA (Figure \ref{fig_chex_celeb}, bottom row) the worst performing group in a balanced scenario is M, by a larger margin than in the previous setting. Consequently, this means a stronger imbalance is required to compensate this effect, but again we find that a value exists (80-20 ratio), which reduces the gap by 3.4 points of accuracy while increasing the overall performance by 0.45\%. Once again, for this fixed task difficulty and balance ratio, ensembles reduce the gap while improving performance for all groups (1.5 and 0.7 points of accuracy for M and F, respectively).

\section{Discussion and conclusions}

In this work, we evaluated the use of simple homogeneous ensembles as a technique to address fairness, as well as the interplay between under-representation of certain groups and per-group task difficulty in this context. 

Importantly, we showcased in synthetic and real datasets that homogeneous ensembles robustly reduce the performance gap between groups while increasing the individual performance of both groups, achieving positive-sum fairness. This departs from a common observation from bias mitigation techniques, which target biases at the expense of a decrease in overall performance, resulting in a leveling-down effect. While generally relevant, this is critical for applications in the healthcare domain, where intentionally reducing the performance of a given sub-population, even if it benefits the whole, may be considered to violate the bioethical principle of non-maleficence. 

Although per-group difficulty can be hard to manipulate in real-life settings, if starting from a balanced or close-to-balanced dataset, the degree of representation could be in principle manipulated during training. Contrary to the established cannon, which advocates for balance as a basic goal when it comes to fairness, we find that the interaction between per-group difficulty and representation shifts the optimal balance point from 50-50. While for similar difficulties the optimal balance sits indeed close to 50\%, as the relative difficulties increase, this number may reach values as high as 80-20. In other words, these results serve as a cautionary tale that standard rebalancing techniques may actually be harmful if the relative difficulty of the task is not taken into consideration.

Taken together, these results highlight the importance of intersectional studies when it comes to the sources of biases, and the potential value of traditional ML tools such as ensembles to achieve positive-sum fairness. Substantial work remains to be done to understand if and how these findings generalize to other ML domains and tasks, opening up a valuable avenue of research.

\section{Limitations}

Although this paper is among the first to explore the fairness implications of deep homogeneous ensembles, particularly in relation to task difficulty and under-representation, it is not without limitations.

First, in the real-world scenarios we analyzed, our notion of task difficulty was limited. We used the performance gap in balanced training datasets as a proxy for task difficulty. Then, based on the assumption that when training samples are equally distributed across different sub-groups, differences in performance can be attributed to task difficulty rather than under-representation. However, this assumption may not hold in cases where sampling is unequal, for example, when samples in one subpopulation exhibit less variability than those in another. Future research should account for this factor when defining task difficulty. Moreover, this may depend on model choice. Ideally, one would want a measure of difficulty which directly captures how often experts would fail for that particular case. Datasets with multiple expert annotators would be greatly beneficial in this regard. We note that this is a similar challenge to the one faced by standard calibration assessment methods, where one does not have access to the true uncertainty of the annotations.

Second, here we experimented with an FC architecture for the synthetic experiments in tabular data, and a CNN architecture for the real scenarios in image classification. More complex architectures could be considered (like Visual Transformers for image classification, for example) to ensure that our conclusions also hold for larger vision models.





\bibliographystyle{ACM-Reference-Format}
\bibliography{sample-base}

\appendix

\section*{Appendix A. Synthetic scenario \textit{rotating decision boundary}}
In this scenario we defined group difficulty by rotating one of the subgroup's decision boundary by a varying rotation angle. This increased difficulty is achieved by using an L1 regularization when training the models, making the rotated subgroup more difficult than the one with a horizontal decision boundary. We follow the same scheme for training models and studying the effect of ensembles as with the \textit{label noise} synthetic scenario.

\begin{figure*}
\centering
\includegraphics[width=1\textwidth]{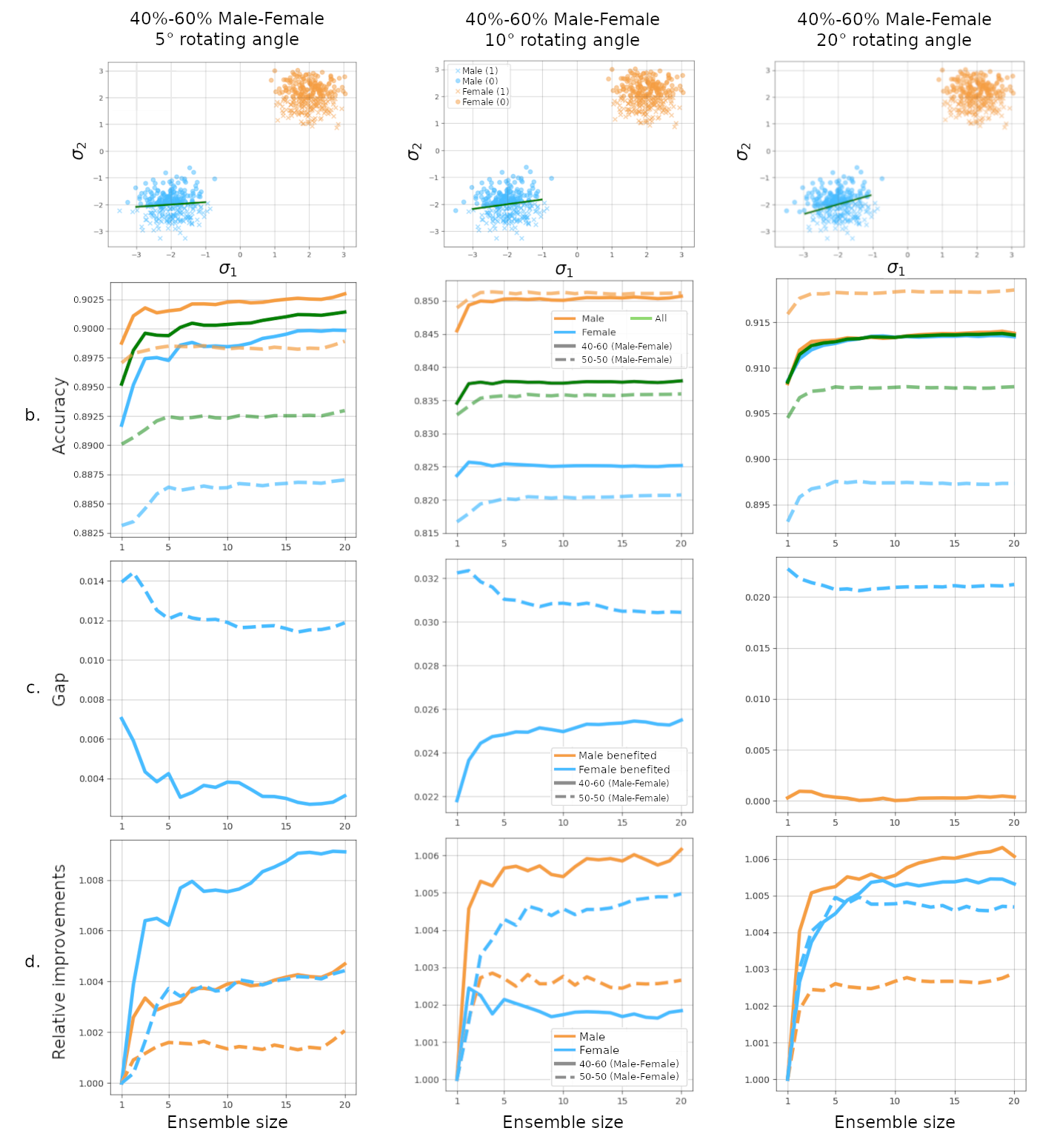}
\caption{Results for the 
 \textit{rotating decision boundary} synthetic scenario. Each column represents a different angle on which the decision boundary is rotated for the female subgroup. The higher the angle, the higher the difficulty for this subgroup. Every case was trained with a sub-representation of males by 40\% (40-60 M-F). (a) Samples of both distributions under different angle-balance configurations (in green, the decision boundary with its corresponding angle). The following rows correspond to (b) model accuracy for male (orange), female (light blue), and overall (green); (c) absolute performance gap (line color represents the group that is being benefited), and (d) relative improvements by the ensemble size. Dashed lines correspond to the same angle configuration but in a balanced scenario (50-50 M-F).}\label{fig_rotating_boundary}
\end{figure*}

Once again we noticed how by using an imbalanced scenario (40-60 male-female in this case) we can achieve a better scenario than for balanced one (50-50 male-female), not only in terms of accuracy for the most difficult group (female), but also in terms of the gap between both groups (see Figure \ref{fig_rotating_boundary} b,c).

\begin{figure*}
\centering
\includegraphics[width=1\textwidth]{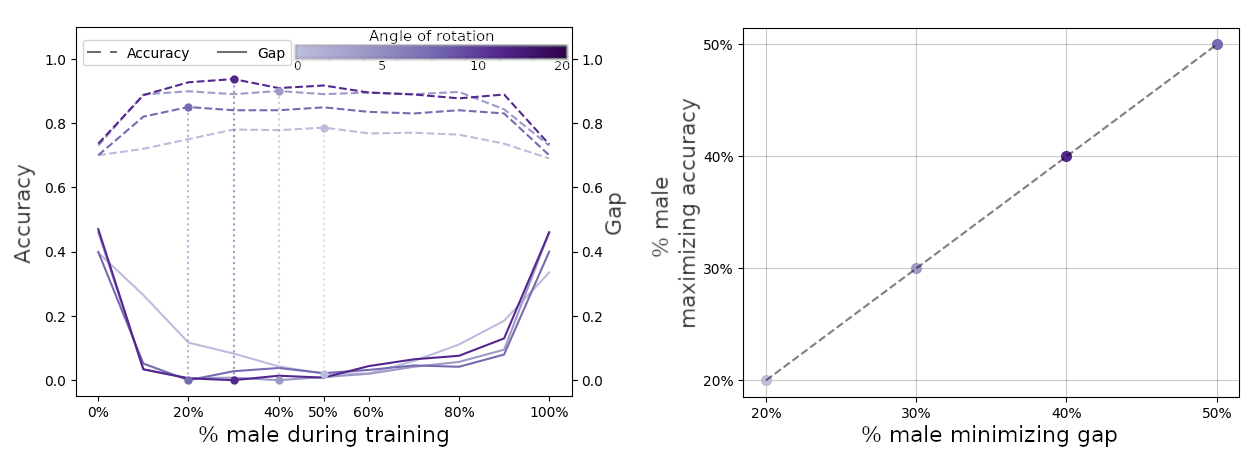}
\caption{(Left) Overall accuracy and gap for different noise percentages by the balance ratio between male and female subgroups for \textit{rotating decision boundary}. (Right) Ideal male-female balance ratio for the different noise percentages. As the discrepancy in difficulty increases for both subgroups, more samples are needed for the most difficult one to achieve the best overall performance and minimum gap between the groups.}
\end{figure*}

\end{document}